\documentclass[10pt, conference, compsocconf]{IEEEtran}

\usepackage[utf8]{inputenc}
\usepackage{amssymb}
\usepackage{array}
\usepackage{float}
\usepackage{graphicx}
\usepackage[colorinlistoftodos]{todonotes}
\usepackage{multirow}
\usepackage{stfloats}
\usepackage{textcomp}
\usepackage{flushend}
\usepackage[hyphens]{url}
\usepackage{amsmath}
\usepackage{chngcntr}
\usepackage{cite}
\counterwithin{paragraph}{subsection}

\newcolumntype{P}[1]{>{\centering\arraybackslash}p{#1}}
\newcolumntype{M}[1]{>{\centering\arraybackslash}m{#1}}

\begin{document}

%
%

\title{Combining Static and Dynamic Features for Multivariate Sequence Classification}


\author{\IEEEauthorblockN{Anna Leontjeva}
\IEEEauthorblockA{Institute of Computer Science\\
University of Tartu\\
Tartu, Estonia\\
Email: anna.leontjeva@ut.ee}
\and
\IEEEauthorblockN{Ilya Kuzovkin}
\IEEEauthorblockA{Computational Neuroscience Lab\\
University of Tartu\\
Tartu, Estonia\\
Email: ilya.kuzovkin@gmail.com}
}

\maketitle

%
%
\begin{abstract}
Model precision in a classification task is highly dependent on the feature space that is used to train the model. Moreover, whether the features are sequential or static will dictate which classification method can be applied as most of the machine learning algorithms are designed to deal with either one or another type of data. In real-life scenarios, however, it is often the case that both static and dynamic features are present, or can be extracted from the data. In this work, we demonstrate how generative models such as Hidden Markov Models (HMM) and Long Short-Term Memory (LSTM) artificial neural networks can be used to extract temporal information from the dynamic data. We explore how the extracted information can be combined with the static features in order to improve the classification performance. We evaluate the existing techniques and suggest a hybrid approach, which outperforms other methods on several public datasets.
\end{abstract}

%
%

%
%
\section{Introduction}
When it comes to a classification task it is quite common to think about two different feature categories: \emph{sequential}, where each data sample is represented by one or many features with their values changing over time (\emph{time-series} or \emph{dynamic} features) and \emph{static} data with each sample described by a set of features, each having a fixed value. Those values do not change in time and are fixed for each sample. We will refer to them as \emph{static} features. 

For almost any sequential dataset it is possible to extract static features out of temporal data. One of the examples of such an approach would be Fourier analysis on EEG signals \cite{knott1942fourier} that transforms signals of arbitrary length to a fixed-size frequency domain. Moreover, in many real-life applications a dataset can already consist of instances that have features of both categories. For example, consider hospital data, where the age and gender of a patient are static features, while heartbeats recorded from the electrodes over some period of time are dynamic features.

Despite the fact that both static and dynamic features may contribute to the classification~\cite{mesnil2014ensemble}, they are rarely used together. One of the reasons is that most machine learning methods are not suitable for processing static and dynamic data simultaneously. Such discriminative algorithms as Random Forest~\cite{breiman2001random}, Support Vector Machine (SVM)~\cite{scholkopf1998support}, feed-forward neural networks~\cite{hornik1989multilayer} take static features as an input. For sequence classification, common methods are variations of Hidden Markov Models (HMM)~\cite{rabiner1989tutorial}, Dynamic Time Warping (DTW)~\cite{ding2008querying} and Recurrent Neural Networks~\cite{hochreiter1991untersuchungen}.
It is also possible to tackle sequential data by transforming sequences into feature vectors that can be fed into a discriminative method. Ensemble \cite{dietterich2000ensemble} methods provide another way to address the issue: predictions made by a \emph{temporal} model on dynamic data are combined with the predictions of a discriminative classifier on static data.

In this paper, we investigate whether there is a better way for extracting useful information from both data modalities to improve the overall classification performance. We devise a data augmentation technique where static features are concatenated with the data representation provided by a dynamic model. We refer to such an approach as \emph{hybrid} and show that the hybrid way of stacking models~\cite{wolpert1992stacked} is in general more beneficial than ensemble methods. We summarize our main contributions as follows:
\begin{itemize}
    
    \item We postulate an idea that combining temporal and static information can boost classification performance.
    
    \item We compare different ways to combine static and dynamic features and propose a hybrid approach that employs an unconventional way of concatenating features.
    
    \item We empirically demonstrate that a hybrid method outperforms ensemble and other baseline methods on several public datasets.
    
    \item We perform a controlled experiment on a synthetic dataset to investigate how dataset characteristics affect the baseline, ensemble and hybrid methods' performance.
    
\end{itemize}

%
%
\section{Preliminaries}
\label{sec:preliminaries}
In this section, we introduce background information on the algorithms and techniques, which will serve as the building blocks for the more complex architectures.
\ \\
\paragraph{Random Forest (RF)} There is a vast amount of discriminative algorithms available. For our experiments we have chosen Random Forest as it provides close to the state of the art performance on many practical problems \cite{scornet2014consistency, denil2013narrowing}. We employ Random Forest in two different ways: as a stand-alone discriminative classifier and as a final predictor, which combines the lower layer features in the ensemble and hybrid architectures. The Scikit-learn \cite{scikit-learn} implementation of Random Forest was used in our experiments.
\ \\
\paragraph{Hidden Markov Models (HMM)} HMM belongs to the class of probabilistic graphical generative approaches, which means they are used to generate samples from a joint distribution of observed and unobserved features. They have been the most frequently used technique for modeling sequence data since about the 80s \cite{Goodfellow-et-al-2016-Book}. Despite many existing variations of HMM, the most commonly used is the 1-order Markov process with discrete hidden states: the probability of a given state depends only on the previous state, while ignoring the rest. If observed values are discrete, HMM is described via three components: an initial distribution of hidden states, a matrix of transition probabilities between the states and a matrix with observation probability distributions for each of the hidden states, which is usually referred to as an \emph{emission matrix}. If observations are not discrete, Gaussian HMMs~\cite{reynolds2015gaussian} are used for the parametrization, so that emission probabilities are described using means and covariance matrix of Gaussian distributions. We use the \texttt{hmmlearn} \cite{hmmlearn} implementation of Gaussian HMM.
\ \\
\paragraph{Long Short-Term Memory (LSTM)} LSTM recurrent neural network \cite{hochreiter1997long} is a subclass of Recurrent Neural Networks family. It has gained popularity by showing the state of the art performance in many fields \cite{graves2013speech, sutskever2014sequence, srivastava2015unsupervised, langkvist2014review}. The core improvement of LSTM over vanilla RNN lies in replacing a usual artificial neuron by a more complicated structure, called an \emph{LSTM unit}. This improvement allows the LSTM network to learn long-term dependencies in the data making the LSTM a perfect fit for the sequential data.

\begin{figure}[h]
    \includegraphics[width=1\linewidth]{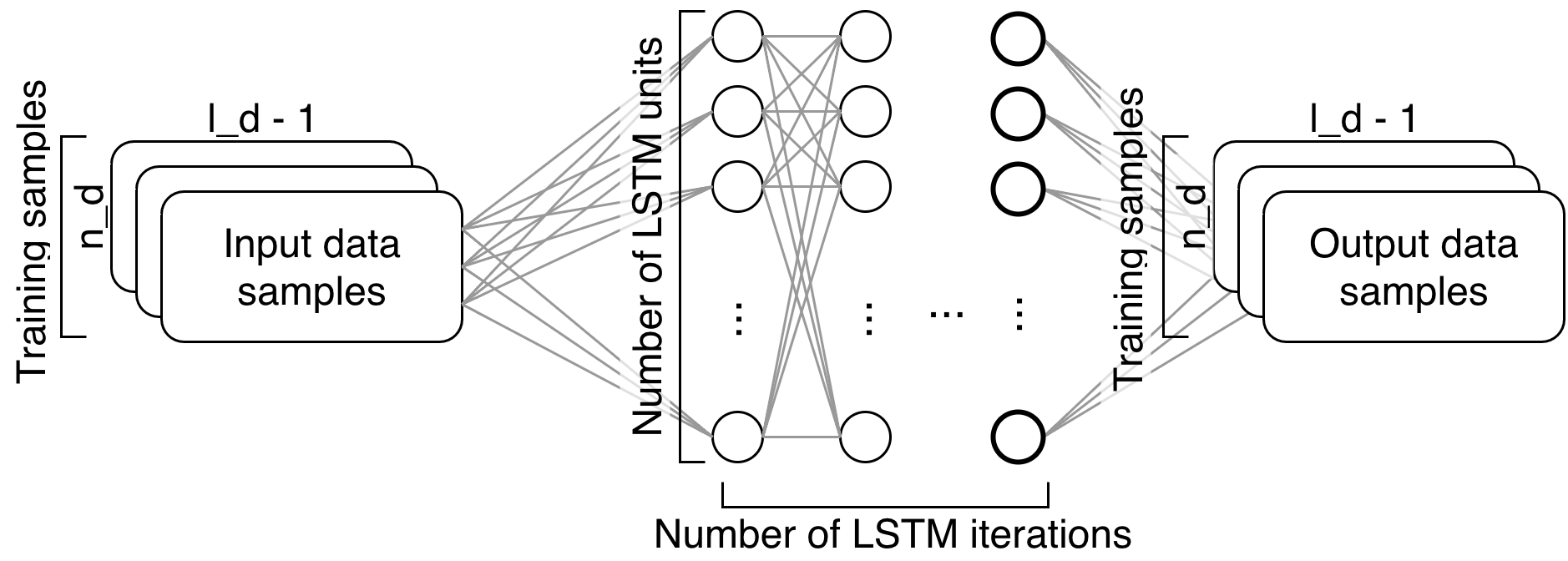}
    \caption{LSTM network architecture for sequence generation. We predict value at time step $t$ from the time steps $1\ldots t-1$. Therefore, the output sequence is the lagged version of the input sequence. The activations of the LSTM units at the last iteration (shown in bold) are used as the features in the $\mathit{HYB_{LSTMA}}\ (12)$ model (see Table \ref{tab:model-list}).}
    \label{fig:lstm-architecture}
\end{figure}

In order to capture temporal dynamics in the data we train a LSTM network to predict each value in the sequence $(x_2, x_3, \dots, x_{l_d})$ given all the previous ones. In the case of multivariate sequences the input size of the network is equal to the number of dynamic features $n_d$; the number of nodes $\mathit{n_{LSTM}}$ in the network is estimated separately for each dataset. We train a single layer of LSTM followed by a fully connected layer of size $\mathit{n_{LSTM}} \times n_d$ (see Figure \ref{fig:lstm-architecture} for the visual explanation).

There are many other possible LSTM architectures, but it is out of the scope of this paper to compare them all. We chose the architecture described above mainly for two reasons: 1) in the current formulation it is more comparable with HMM-based generative models and, 2) among other architectures we tried, this one provided best or comparable results.

In all LSTM models we use mean squared error (MSE) between the true test sequence and the generated sequence as our error function, RMSProp \cite{dauphin2015rmsprop} acts as the optimization method. The Keras Deep Learning library \cite{keras} provides the implementation of LSTM.
\ \\
\paragraph{Time Spatialization} To be able to apply Random Forest on dynamic data we need to transform the raw data into an appropriate feature space. Let $n_d$ be the number of sequential features in a data sample, each of them of length $l_d$. In this case, every sample can be represented by a matrix of size $n_d \times l_d$:

\ \\
{\centering
$\begin{bmatrix}
    x_{11} & x_{12} & \dots  & x_{1l_d} \\
    x_{21} & x_{22} & \dots  & x_{2l_d} \\
    \vdots & \vdots & \ddots & \vdots \\
    x_{n_d1} & x_{n_d2} & \dots  & x_{n_dl_d}
\end{bmatrix}$\\}
\ \\

By flattening the matrix we obtain a feature vector of length $n_d \cdot l_d$ thus transforming a data sample from the time domain to the feature space: 
\small
$$(x_{11}, x_{12}, \dots, x_{1l_d}, x_{21}, x_{22}, \dots, x_{2l_d}, \dots, x_{n_d1}, x_{n_d2}, \dots, x_{n_dl_d}).$$
\normalsize
Data represented in such a form can be used to train Random Forest classifier \cite{baydogan2015learning}.
\ \\
\paragraph{Generative Models for Classification} Some of the approaches that we consider use a generative model (HMM or LSTM) as a discriminative classifier. We consider the binary classification task, however, it can be easily extended to multiple classes. First, we split the training set into two subsets: a subset of all samples with a positive class label and another subset of all samples with a negative class label. Next, we train two generative models: $G_\texttt{POS}$ on the positive samples and $G_\texttt{NEG}$ on the negative samples. In order to classify a new sample $s$, we compute the log-likelihood of $L(s | G_\texttt{POS})$ and $L(s | G_\texttt{NEG})$, afterwards we use a binary decision rule
$$L(s | G_\texttt{POS}) \geq L(s | G_\texttt{NEG})$$
to assign the class label to the sample $s$. 

An important benefit of a generative model is the possibility to use sequences of a varying length. Discriminative methods, such as Random Forest, are not applicable in such cases, thus making a temporal model the only suitable approach. In our experiments we use only datasets with the sequences of fixed length for the compatibility reasons. It is straightforward to extend the code for sequences of a varying length. 
\ \\
\paragraph{Log-likelihoods and Ratios as Features} Once we estimate log-likelihood for a sample $s$, as described in the previous paragraph, we can use $L(s | G_\texttt{POS})$ and $L(s | G_\texttt{NEG})$ estimations as features. This is the approach we take in ensemble methods $\mathit{ENS_{HMM}}$ and $\mathit{ENS_{LSTM}}$. These two additional features indicate how likely it is that the sample $s$ is generated by the positive generative model and what is the likelihood of it being generated by the negative generative model.

Another derivative feature is the ratio of those two measures. In case of HMM it is
$$r_{\texttt{HMM}} = L(s | G_\texttt{POS}) - L(s | G_\texttt{NEG})$$
and for LSTM
$$r_{\texttt{LSTM}} = \log(\frac{\mathit{MSE}_{\texttt{NEG}}}{\mathit{MSE}_{\texttt{POS}}})$$
where $\mathit{MSE}_{\texttt{POS}}$ and $\mathit{MSE}_{\texttt{NEG}}$ are the mean squared errors between the true output sequence and the generated sequence. Our experiments have shown that combining static features with ratios yields better performance than combining static features with raw log-probabilities or MSE scores.

%
%
\section{Methods}
\label{sec:methods}
In this section, we first describe the approaches that can be thought of as competitors to the hybrid models, then we explain our main contribution --- the concept of a hybrid model and its versions.

Subsection~\ref{sec:single-on-uni} describes the methods that use only static or only dynamic features (\emph{unimodal}) with a single discriminative algorithm. Subsection~\ref{sec:single-on-bi} makes a step forward by using a feature space that includes both data modalities (\emph{bimodal}) transformed to make the dataset suitable for the algorithm at hand. For example, if we apply Random Forest to sequential data we employ spatialization; if we need to apply a temporal model to static data we transform its feature values into ``fake" sequences: let the value of a static feature $f_i$ of a sample be $x$, $f_i$ will be represented by a sequence of length $l_d$ with value $x$ at each time step. Subsection~\ref{sec:multi-on-bi} discusses what are the options to construct feature spaces that combine static and dynamic features effectively and describes ensemble and hybrid methods applied to these feature spaces. 

%
%
\subsection{Stand-alone Models on Unimodal Data}
\label{sec:single-on-uni}
In this section we describe the simplest baseline approaches. These are the models that are fit for only one data modality: either static or dynamic.
\ \\
\paragraph{Random Forest on Static or Dynamic Features} The most straightforward way to handle a multimodal dataset is to build a model on static features only. In this work such an approach is referred to as $\mathit{RF_{s}}\ (1)$\footnote{The number in brackets stands for the model ID we use throughout the text.}. We denote the number of static features in a dataset as $n_s$. If the dynamic data has the sequences of a fixed length, then it is straightforward to apply Random Forest on this data --- we use time spatialization to represent time series as one long feature vector. One obvious drawback of such an approach is low performance due to the curse of dimensionality when sequences are extremely long \cite{keogh2011curse}. We use this baseline to estimate if the use of a temporal model is justified. Our intuition is that if Random Forest is able to achieve the same performance on the dynamic data as temporal models do, then the particular dynamic dataset does not have a strong temporal component. We denote this method as $\mathit{RF_{d}}\ (4)$ and use it for the comparison with temporal models such as HMM and LSTM. 
\ \\
\paragraph{Hidden Markov Models on Dynamic Features} The method denoted as $\mathit{HMM_d}\ (2)$ is a direct application of HMM to sequential data. We use HMM as a classifier as described in the paragraph ``Generative Models for Classification" of Section~\ref{sec:preliminaries}.
\ \\
\paragraph{Long Short-Term Memory on Dynamic Features} The method denoted as $\mathit{LSTM_d}\ (3)$ uses the same technique as $\mathit{HMM_d}\ (2)$ to act as a classifier. The architecture of both $G_{\texttt{POS}}$ and $G_\texttt{NEG}$ networks are shown in Figure \ref{fig:lstm-architecture}. 
%
%
\subsection{Stand-alone Models on Bimodal Data}
\label{sec:single-on-bi}
The second class of baseline approaches utilizes both static and dynamic data by concatenating them in such a way that a single classification method is applicable. This is the most na\"ive way of using both data modalities simultaneously, and, as it will be discussed later, has obvious limitations.
\ \\
\paragraph{Random Forest on Static and Dynamic Features} The method under the name $\mathit{RF_{s,d}}\ (5)$ transforms dynamic data to static, concatenates it with the original static features and employs Random Forest on the resulting feature set.
\ \\
\paragraph{Hidden Markov Model on Static and Dynamic} The method implemented in $\mathit{HMM_{s,d}}\ (6)$ transforms static features into ``fake" sequences: for the dataset with $n_d$ dynamic features of length $l_d$ it produces $n_s$ additional dynamic features of length $l_d$. All of the values along these sequences are constant and equal to the original value of the static feature. Using this trick we extend dynamic feature set from $n_d$ features to $n_d + n_s$ features and apply HMM on it similarly to $\mathit{HMM_d}\ (2)$.
\ \\
\paragraph{Long Short-Term Memory on Static and Dynamic Features} Using the trick from $\mathit{HMM_{s,d}}\ (6)$ we obtain the dynamic features from the static features, concatenate them with the original dynamic features and train an LSTM classifier on the combined feature set. The learning algorithm itself is analogous to $\mathit{LSTM_d}\ (3)$. We refer to this approach as $\mathit{LSTM_{s,d}}\ (7)$.

%
%
\subsection{Multiple Models on Bimodal Data}
\label{sec:multi-on-bi}

\begin{table*}[bp]
\caption{List of models and the feature sets they operate upon.}
\label{tab:model-list}
\centering
\begin{tabular}{lll|cc|ccc|cc|c|}
\cline{4-11}
\multicolumn{1}{c}{}                            &                          & \multicolumn{1}{c|}{}           & \multicolumn{2}{c|}{Raw features} & \multicolumn{3}{c|}{Predictions}     & \multicolumn{2}{c|}{Ratios} & Activations \\ \cline{2-11} 
\multicolumn{1}{c|}{}                           & \multicolumn{1}{l|}{Nr.} & \multicolumn{1}{c|}{Model name} & Static          & Dynamic         & RF         & HMM        & LSTM       & HMM          & LSTM         & LSTM        \\ \hline
\multicolumn{1}{|l|}{\multirow{7}{*}{Stand-alone}}   & \multicolumn{1}{l|}{1}   & $\mathit{RF_{s}}$                        & \checkmark      &                 &            &            &            &              &              &             \\
\multicolumn{1}{|l|}{}                          & \multicolumn{1}{l|}{2}   & $\mathit{HMM_{d}}$                       &                 & \checkmark      &            &            &            &              &              &             \\
\multicolumn{1}{|l|}{}                          & \multicolumn{1}{l|}{3}   & $\mathit{LSTM_{d}}$                      &                 & \checkmark      &            &            &            &              &              &             \\
\multicolumn{1}{|l|}{}                          & \multicolumn{1}{l|}{4}   & $\mathit{RF_{d}}$                        &                 & \checkmark      &            &            &            &              &              &             \\
\multicolumn{1}{|l|}{}                          & \multicolumn{1}{l|}{5}   & $\mathit{RF_{s,d}}$                      & \checkmark      & \checkmark      &            &            &            &              &              &             \\
\multicolumn{1}{|l|}{}                          & \multicolumn{1}{l|}{6}   & $\mathit{HMM_{s,d}}$                     & \checkmark      & \checkmark      &            &            &            &              &              &             \\
\multicolumn{1}{|l|}{}                          & \multicolumn{1}{l|}{7}   & $\mathit{LSTM_{s,d}}$                    & \checkmark      & \checkmark      &            &            &            &              &              &             \\ \hline
\multicolumn{1}{|l|}{\multirow{2}{*}{Ensemble}} & \multicolumn{1}{l|}{8}   & $\mathit{ENS_{HMM}}$                     &                 &                 & \checkmark & \checkmark &            &              &              &             \\
\multicolumn{1}{|l|}{}                          & \multicolumn{1}{l|}{9}   & $\mathit{ENS_{LSTM}}$                    &                 &                 & \checkmark &            & \checkmark &              &              &             \\ \hline
\multicolumn{1}{|l|}{\multirow{3}{*}{Hybrid}}   & \multicolumn{1}{l|}{10}  & $\mathit{HYB_{HMM}}$                     & \checkmark      &                 &            &            &            & \checkmark   &              &             \\
\multicolumn{1}{|l|}{}                          & \multicolumn{1}{l|}{11}  & $\mathit{HYB_{LSTM}}$                    & \checkmark      &                 &            &            &            &              & \checkmark   &             \\
\multicolumn{1}{|l|}{}                          & \multicolumn{1}{l|}{12}  & $\mathit{HYB_{LSTMA}}$                   & \checkmark      &                 &            &            &            &              &              & \checkmark  \\ \hline
\end{tabular}
\end{table*}

Dealing with bimodal data is the main focus of this work. In this section we look into different ways of combining static and dynamic features.
\ \\
\subsubsection{Ensemble Models}
\label{sec:ensemble-models}
With an ensemble approach one can train different models for different data modalities (for example, Random Forest for static features and LSTM for dynamic features) and combine their predictions using a linear model or another layer of Random Forest (or any other discriminative method). 

In this work we have two methods based on the ensemble approach: $\mathit{ENS_{HMM}}\ (8)$, which takes the predictions made by Random Forest and the predictions made by HMM classifier, and the $\mathit{ENS_{LSTM}}\ (9)$ that combines Random Forest predictions with the predictions of LSTM classifier. In both these models the final prediction is obtained by training an additional Random Forest model using predictions as features.

\begin{figure}[h]
    \includegraphics[width=1\linewidth]{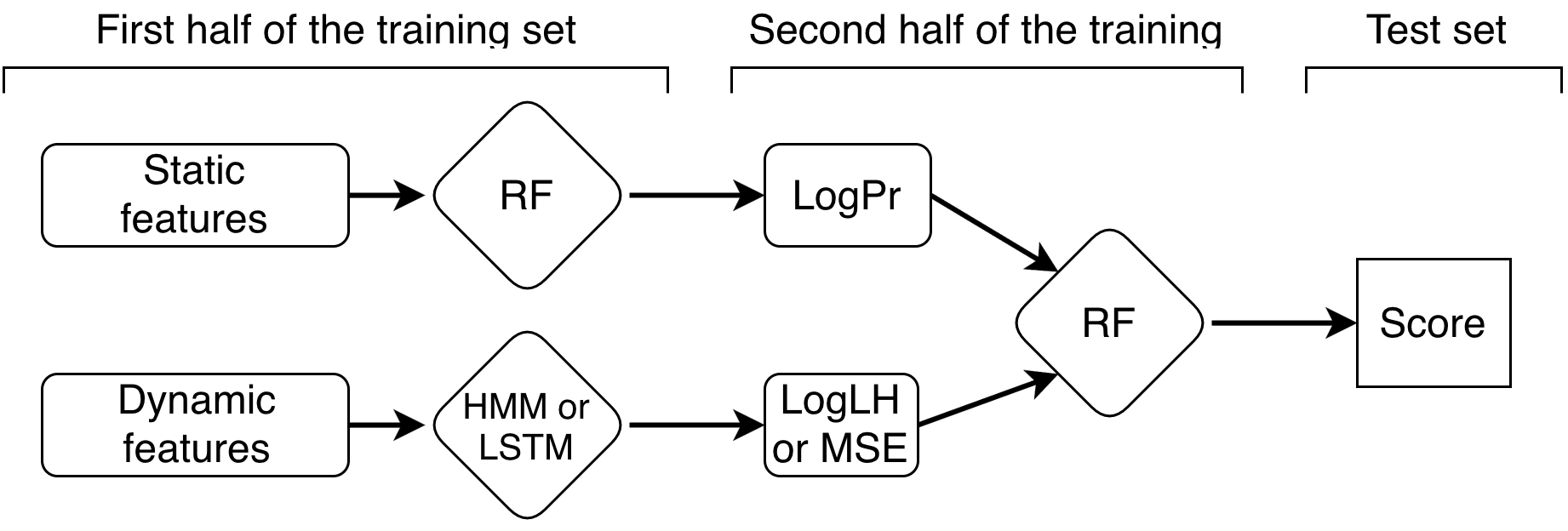}
    \caption{Architecture of an ensemble. The first half of the training set is used to create models according to the data modality: Random Forest for static data and a generative model for the dynamic data. These models are applied to the second half of the training set and to the test set to extract predictions and form a new feature space. Random Forest is trained on the enriched second half of the training set and evaluated on the enriched test set.}
    \label{fig:ensemble-architecture}
\end{figure}

\paragraph{Ensemble of HMM and RF} The ensemble method $\mathit{ENS_{HMM}}\ (8)$ has two stages. First stages works with the first half of the training set --- Random Forest is trained on the static features and HMM classifier on the dynamic ones. In the second stage we use the samples from the second half of the training set and estimate class probabilities for each sample using the models trained in the first stage. In case of a binary classification problem each sample is represented by 4 features. We obtain a new dataset where each sample $s$ from the second half of the original training set is represented by: log probabilities that $s \in \texttt{POS}$ and $s \in \texttt{NEG}$ provided by Random Forest, and $L(s | G_\texttt{POS})$, $L(s | G_\texttt{NEG})$ log-likelihoods provided by HMM. In the similar way we feed samples from the original test set into these models and obtain a test set with the same 4-dimensional feature space. Finally, we train Random Forest on 4-dimensional training set and evaluate it on the corresponding test set. For the detailed explanation of the experimental pipeline see Section~\ref{sec:pipeline}.
\ \\
\paragraph{Ensemble of LSTM and RF} LSTM ensemble $\mathit{ENS_{LSTM}}\ (9)$ builds a new feature set in a similar to $\mathit{ENS_{HMM}}\ (8)$ way. The first two features are the same as in the case of $\mathit{ENS_{HMM}}\ (8)$. The second two features are obtained with $\texttt{POS}$ and $\texttt{NEG}$ LSTM networks. Namely, an input sequence of a data sample $s$ is fed into each of the networks, and the corresponding output sequences are generated. Per class mean squared errors $\mathit{MSE}_{\texttt{POS}}$ and $\mathit{MSE}_{\texttt{NEG}}$ between the true output sequence and the generated sequences are used as the new features for the sample $s$. The number of such features is equal to the total number of classes.

%
%
\ \\
\subsubsection{Hybrid Models}
\label{sec:hybrid-models}

\begin{table*}[bp]
\caption{Descriptions of the real-life datasets.}
\label{table:datasets}
\centering
\begin{tabular}{l|ccccccc}
\multicolumn{1}{c|}{Datasets} & \multicolumn{1}{c}{Samples} & \multicolumn{1}{c}{Train set} & \multicolumn{1}{c}{Test set} & \multicolumn{1}{c}{\begin{tabular}[c]{@{}c@{}}Static\\ features\end{tabular}} & \multicolumn{1}{c}{\begin{tabular}[c]{@{}c@{}}Dynamic\\ features\end{tabular}} & \multicolumn{1}{c}{\begin{tabular}[c]{@{}c@{}}Sequence \\ length\end{tabular}} & \multicolumn{1}{c}{\begin{tabular}[c]{@{}c@{}}Source of\\ benchmark\end{tabular}} \\ \hline
ECoG & 10584 & \multicolumn{2}{l}{5-fold CV} & 320 & 64 & 300 & \cite{ECoGbest} \\
FordA & 4291 & 1320 & 3601 & 500 & 1 & 500 & \cite{bagnall2012transformation} \\
FordB & 4446 & 810 & 3636 & 500 & 1 & 500 & \cite{bagnall2012transformation} \\
Phalanges & 2658 & 1800 & 858 & 80 & 1 & 80 & \cite{UCRArchive} \\
Yoga & 3300 & 300 & 3000 & 426 & 1 & 426 & \cite{baydogan2015learning}
\end{tabular}
\end{table*}

The general idea of the hybrid approach is to employ generative models such as HMM or LSTM to act as feature extractors from dynamic data. As generative models are able to generate sequences from the training data distribution, it is reasonable to assume that these models can capture temporal dynamics in the data. Therefore, the features extracted using these models can act as an approximation for temporal information contained in the data. These features are concatenated with the static features and a discriminative classifier (Random Forest) is used to build the final predictor.

Since na\"ive ways of combining dynamic data with static features give poor performance (see Figure \ref{fig:acc_datasets}) we use the data representation provided by temporal models to obtain a fixed-size feature set that contains knowledge extracted from the temporal component of dynamic data.

There are different features that can be extracted from generative models, with one such example being the Fisher kernels~\cite{jaakkola1999exploiting}. In our experiments we use log-likelihood ratios, MSE ratios or LSTM activations as features for the enrichment of the static feature set. In the following subsections we go through the hybrid architectures we have explored.

\begin{figure}[h]
    \includegraphics[width=1\linewidth]{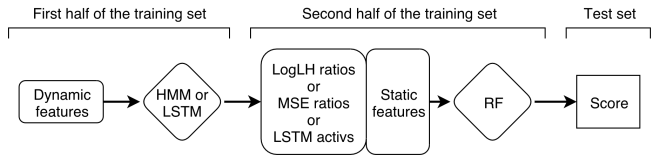}
    \caption{Architecture of the hybrid model. First half of the training set is used to create a model which will act as a \emph{feature extractor}. Feature extractor is applied to \emph{enrich} the feature set of the second half of the training set and the test set with additional features. Random Forest is trained on the second half of the training set to create a final classifier, which is evaluated on the test set.}
    \label{fig:hybrid-architecture}
\end{figure}

\paragraph{Hybrid of Static Features and HMM Ratios} In the $\mathit{HYB_{HMM}}\ (10)$ method two generative HMM models are built on the first half of the training set: one for the samples with the positive class labels and one for the samples with the negative class labels. These models are used to enrich both static features of the second half of the training set with $r_{\texttt{HMM}}$ ratios as well as the static features of the test set. Finally, Random Forest is trained on the enriched feature set and evaluated on the enriched test set.
\ \\
\paragraph{Hybrid of Static Features and LSTM Ratios} In a very similar fashion we can use LSTM to build generative models. $\mathit{HYB_{LSTM}}\ (11)$ extracts MSE ratios $r_{\texttt{LSTM}}$ and enriches the set of static features on the second half of the training set. Random Forest is trained on the enriched training set and evaluated on the enriched test set.
\ \\
\paragraph{Hybrid of Static Features and LSTM Activations} Depending on how detailed information we want to give to the final classifier we can choose which features to extract from a generative model. The log-likelihood ratios are almost the most compressed form of the information about the data samples. Less compressed features depend on the inner workings of a particular generative model. For a LSTM network we can take activations of the last LSTM layer at the last iteration and use those activations to enrich the set of static features. The model that does that is denoted as $\mathit{HYB_{LSTMA}}\ (12)$. The final step is similar to all other hybrid architectures: Random Forest is trained on the enriched training set (in this case it consists of static features concatenated with LSTM activations) and evaluates the performance on the test set. In our experiments, however, such an approach is less accurate than the one based on the log-likelihoods.

In Table \ref{tab:model-list} we summarize all the models and mark the feature sets used by each model. Note that in case of univariate datasets some of the models are virtually the same: model (5) is the same as the model (1), model (7) as the same as the model (3) and the model (6) is the same as the model (2). The reason is that for univariate datasets we use spatialized dynamic data as static features.

%
%
\section{Evaluation}
In this section we explain the experimental setup and the model performance estimation process.

\subsection{Experimental Pipeline}
\label{sec:pipeline}

In the case of hybrid methods the splitting strategy and feature handling are not straightforward. In this subsection we explain the full pipeline both in the case of a training-test split and in the case of cross-validation.
\ \\
\paragraph{Training / test split} Splitting the data entails non-trivial steps in order to perform feature enrichment without introducing bias into the models. Methods (1)--(7) that do not make use of feature enrichment employ the standard machine learning pipeline where a training set is divided into training and validation subsets. Model hyperparameters are estimated on the validation set and the final model is trained on the whole training data and tested on the test data.

In case of ensembles and hybrids, however, the pipeline differs. First, the training set is divided into two equal halves, let us call them $\text{training}_A$ and $\text{training}_B$. $\text{Training}_A$ is used to train the first tier of models, which we call \emph{feature extractors} --- given a data sample their purpose is to output additional features that describe that sample. In case of ensembles these features are log-likelihoods of class labels as discussed in subsection~\ref{sec:ensemble-models}, for hybrids the extracted features can take any of the forms discussed in \ref{sec:hybrid-models}. Hyperparameters of the feature extractor models are estimated on a validation subset of $\text{training}_A$.

Once feature extractors are fully trained we use them to enrich $\text{training}_B$ and test sets: dynamic data of each sample from those sets is fed into a feature extractor. The extracted features are concatenated with the static features of that data sample. For example in case of $\mathit{HYB_{LSTMA}}\ (12)$, if we had a dataset with $n_d$ dynamic features and $n_s$ static features, and we use LSTM network with 128 LSTM units as the feature extractor, then the new feature space would be of size $n_s + 128$.

The final step of the pipeline is to train a second tier --- a discriminative model (Random Forest) --- on the new feature space. Hyperparameters are estimated on the validation subset of enriched $\text{training}_B$, after that a final model is trained on the whole enriched $\text{training}_B$ set and tested on the enriched test set.

The complete process is depicted in Figures \ref{fig:ensemble-architecture} and \ref{fig:hybrid-architecture}.
\ \\
\paragraph{Cross-validation} In the case of proposed hybrid architecture, cross-validation allows for a more efficient use of data.

Instead of dividing the data set into two parts as it was done for the training/test split case, we split data set into $\{C_1, \ldots, C_k\}$ chunks, where $k$ is the number of cross-validation folds. We train a feature extractor model using the data samples from chunks $\{C_1, \ldots, C_{i-1}, C_{i+1}, C_k\}$ and apply that model to enrich the samples from the $C_i$ chunk. This process is repeated for every chunk and as a result the whole dataset is enriched without introducing overfitting bias. 

The next iteration is to train a second tier model (Random Forest) on the enriched data. This is once again done using cross-validation and exactly on the same chunks of data we used before. The process is no different from the classical application of cross-validation: a model is trained on chunks $\{C_1, \ldots, C_{i-1}, C_{i+1}, C_k\}$ and evaluated on the $C_i$ chunk. The reported accuracy is the average accuracy over $k$ folds.
\ \\
\paragraph{Hyperparameter Optimization} In order to find the best hyperparameter combination for a model, we apply Spearmint \cite{snoek2012practical} to search through the parameter space. The method behind the tool is Bayesian optimization and it has been shown to be able to find hyperparameters that yield performance equal or superior to that achieved using other hyperparameter optimization techniques. In our experiments every dataset has its own set of parameters, see Table \ref{tab:hyperparameters} for the details.

\begin{table}[h]
\caption{Estimated hyperparameters. Column names stand for: LSTM Size, Dropout, Optimization method, Batch size, number of Epochs; number of HMM States, number of Iterations; number of RF Trees.}
\label{tab:hyperparameters}
\small
\centering
\begin{tabular}{l|ccccc|cc|c}
          & \multicolumn{5}{c|}{LSTM}      & \multicolumn{2}{c|}{HMM} & RF  \\ \hline
Dataset   & S    & D   & O       & B  & E  & S          & I           & T   \\ \hline
ECoG      & 2000 & 0.5 & rmsprop & 32 & 50 & 6          & 50          & 500 \\
FordA     & 512  & 0.0 & rmsprop & 1  & 20 & 2          & 50          & 500 \\
FordB     & 512  & 0.0 & rmsprop & 1  & 20 & 2          & 50          & 500 \\
Phalanges & 128  & 0.0 & rmsprop & 1  & 10 & 2          & 50          & 500 \\
Yoga      & 256  & 0.0 & rmsprop & 1  & 10 & 2          & 50          & 500
\end{tabular}
\end{table}

%
%
\subsection{Datasets}
\label{sec:datasets}
We compare the described approaches on several datasets from different domains as well as on a simulated data. In this section we describe the datasets and their properties.
\ \\
\paragraph{Synthetic ARMA Dataset} In order to be able to compare results with the ground truth and form the intuition how much information can be extracted from different types of features, we have generated a synthetic dataset with specific properties. Namely, one of the posed questions of this work is whether combining static and dynamic features can boost the overall performance on a given dataset. We model the required conditions by splitting the data into four blocks in the way explained in Table \ref{tab:fourblocks}. Block 1 has samples with positive labels and values for the static features are generated from $G_\texttt{POS}$ model, while the dynamic features are generated from the $G_\texttt{NEG}$ model. In block 2 the situation is reversed. The two last blocks have correct labels for both parts. Therefore, models that do not use information from both sources should be in a worse position. Indeed, models (1)--(4) cannot achieve accuracy of more than $0.75$ as can be seen in Figure \ref{fig:synthetic-performance-all}.

\begin{table}[h]
\small
\renewcommand{\arraystretch}{2}
\caption{Synthetic dataset is designed in a specific way. Each block contains static and dynamic data, however the dynamic data in block 1 and static data in block 2 are useless, making it impossible for a model that operates only on one data modality to classify the whole dataset correctly.}
\label{tab:fourblocks}
\centering
\begin{tabular}{m{0.06\linewidth} 
>{}m{0.35\linewidth}
>{\centering}m{0.08\linewidth} >{\centering}m{0.12\linewidth}>{\centering\arraybackslash}m{0.17\linewidth}}
    Block & Classifiability & Label & Static & Dynamic \\ \hline
        
    1 & Classifiable by discriminative model as \texttt{T}, but generative model will confuse it for \texttt{F} &
    \texttt{T} &
    {\centering$\sim\mathcal{N}_\texttt{T}$} &
    {\centering$\sim \text{ARMA}_{\texttt{F}}$} \\
    
    2 & Classifiable by generative model as \texttt{T}, but discriminative model will confuse it for \texttt{F} &
    \texttt{T} &
    $\sim\mathcal{N}_\texttt{F}$ &
    $\sim \text{ARMA}_{\texttt{T}}$ \\
        
    3, 4 & Classifiable by both as \texttt{F} &
    \texttt{F} &
    $\sim\mathcal{N}_{\texttt{F}}$ &
    $\sim \text{ARMA}_{\texttt{F}}$ \\
\end{tabular}
\end{table}

The dynamic features are simulated with $\text{ARMA}(p, q)$ process \cite{hamilton1994time}, where the orders of autoregressive (AR) and moving average (MA) parts are drawn from a uniform distribution, $p, q \in \{1,\ldots,5\}$, while coefficients of AR and MA processes are $\alpha_i, \beta_i\sim \mathcal{U}(-0.1, 0.1)$, respectively.
All values of the static features in the synthetic dataset are drawn from the Gaussian distribution, $\mathcal{N}(\mu, \sigma^{2})$, where $\mu\sim\mathcal{U}(0,2)$ and $\sigma\sim\mathcal{U}(0,2)$. 

\begin{figure}[h]
    \centering
    \includegraphics[width=1\linewidth]{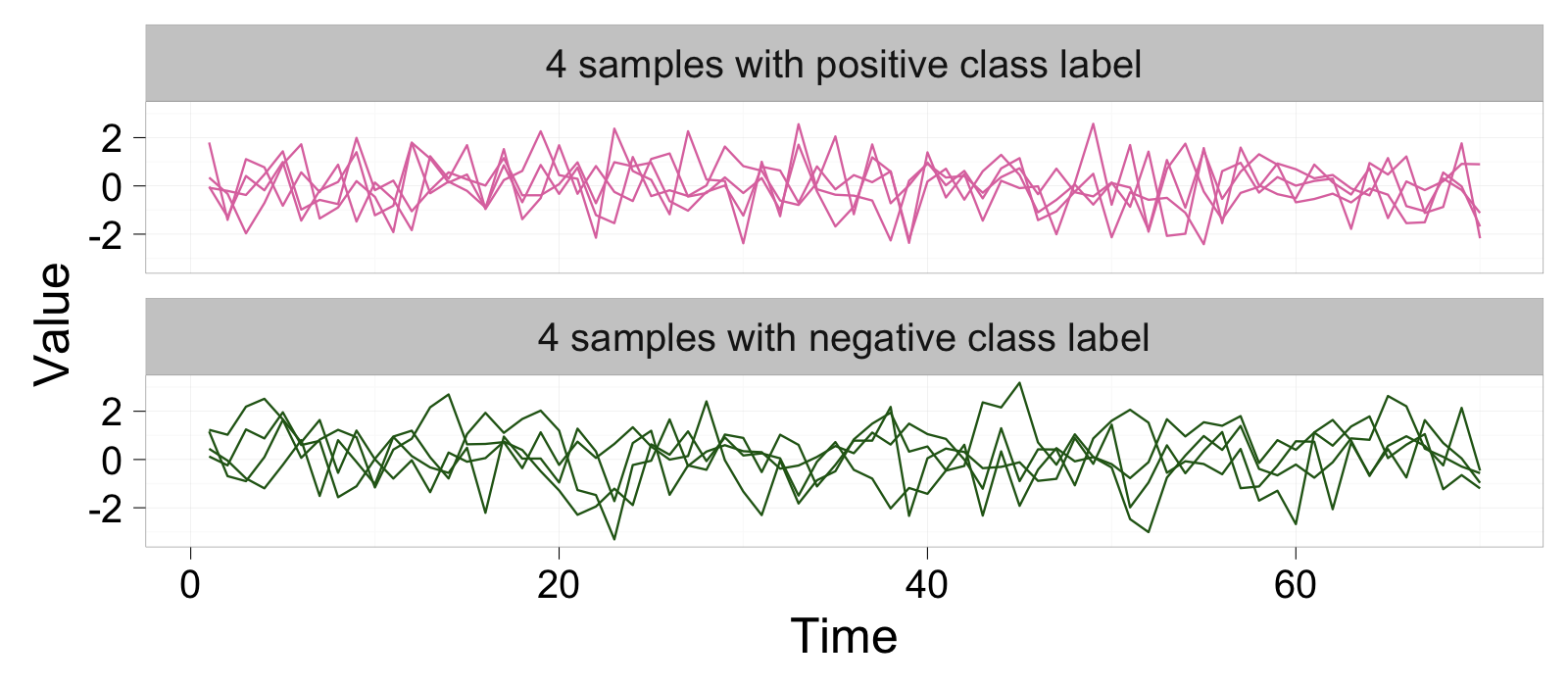}
    \caption{Example of generated synthetic timeseries.}
    \label{fig:synthetic-raw-ts}
\end{figure}

For illustration purposes Figure \ref{fig:synthetic-raw-ts} depicts eight randomly chosen timeseries from the created synthetic dataset. The difference between classes is not obvious, and, therefore, is not overly simple as a classification task.

\begin{figure*}[tp]
    \centering
    \includegraphics[width=1\linewidth]{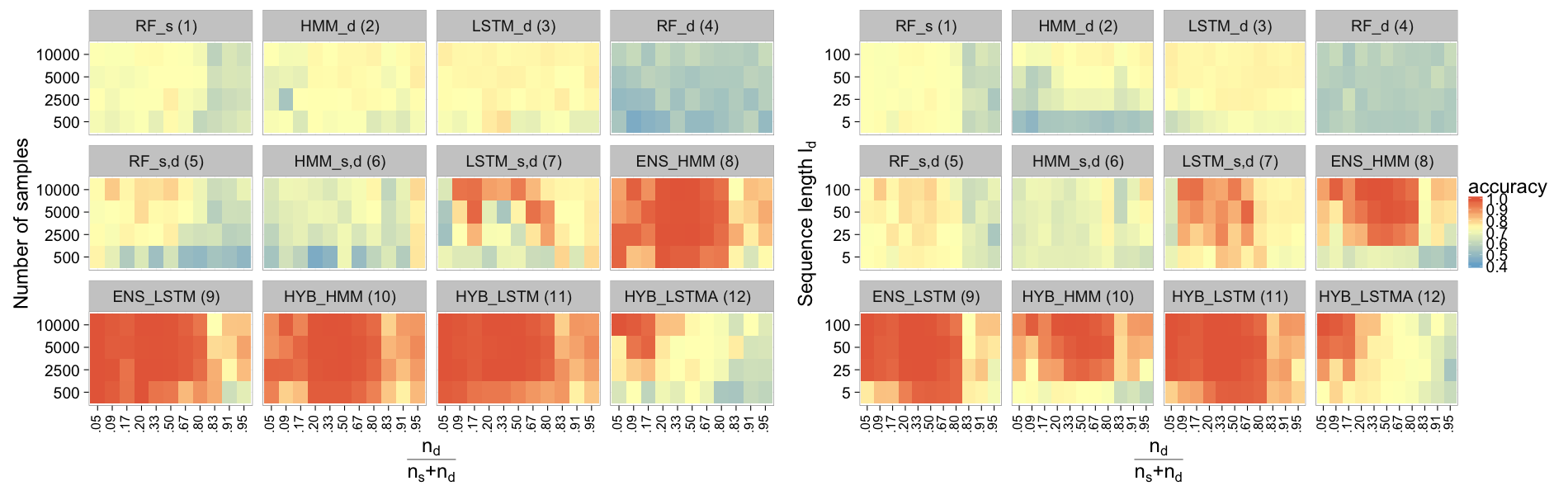}
    \caption{Performance of 12 models on the synthetic dataset with varying parameters. The ratio between the number of dynamic features and the total number of features plotted on the horizontal axis. The vertical axis corresponds to the size of the dataset on the left plot and to the length of the sequence on the right plot.}
    \label{fig:heatmap}
\end{figure*}

\paragraph{Real-life datasets} We use datasets with different aspects: few univariate time-series widely used in the literature --- FordA and FordB \cite{UCRArchive} and a multivariate dataset from a particular domain --- classification of electrocorticography (ECoG) recordings from BCI competition III \cite{lal2004methods}. Also, we show the results on the Phalanges and Yoga datasets \cite{UCRArchive}, where the baseline methods perform as well as ensemble and hybrid approaches, and we discuss why it is the case. For the characteristics of the chosen datasets the reader is referred to Table \ref{table:datasets}. 

Benchmarks from the literature exist for all of the datasets except for the ECoG dataset. To the best of our knowledge we compare our scores with the highest reported results and follow the same data splitting strategies. Namely, we demonstrate the results on the train and test sets of the same size as used in the literature. The sources of the benchmarks are provided in Table \ref{table:datasets}. It is worth mentioning that despite the fact that the best result gained for the ECoG dataset during the BCI Competition has accuracy of $0.91$ \cite{ECoGbest}, the authors use elaborate hand-crafted features extraction methods such as combination of bandpower, CSSD/Waveform Mean and Fisher Discriminant Analysis. Since we do not have access to the features they have used, we limit ourselves to classical Fourier analysis of ECoG signal. Due to the differences in preprocessing the fair comparison cannot be easily drawn. For ECoG dataset we apply 5-fold cross-validation and compute the mean of accuracies over the folds.

%
%
\subsection{Results}
\label{sec:results}

\begin{figure}[h]
    \centering
    \includegraphics[width=1\linewidth]{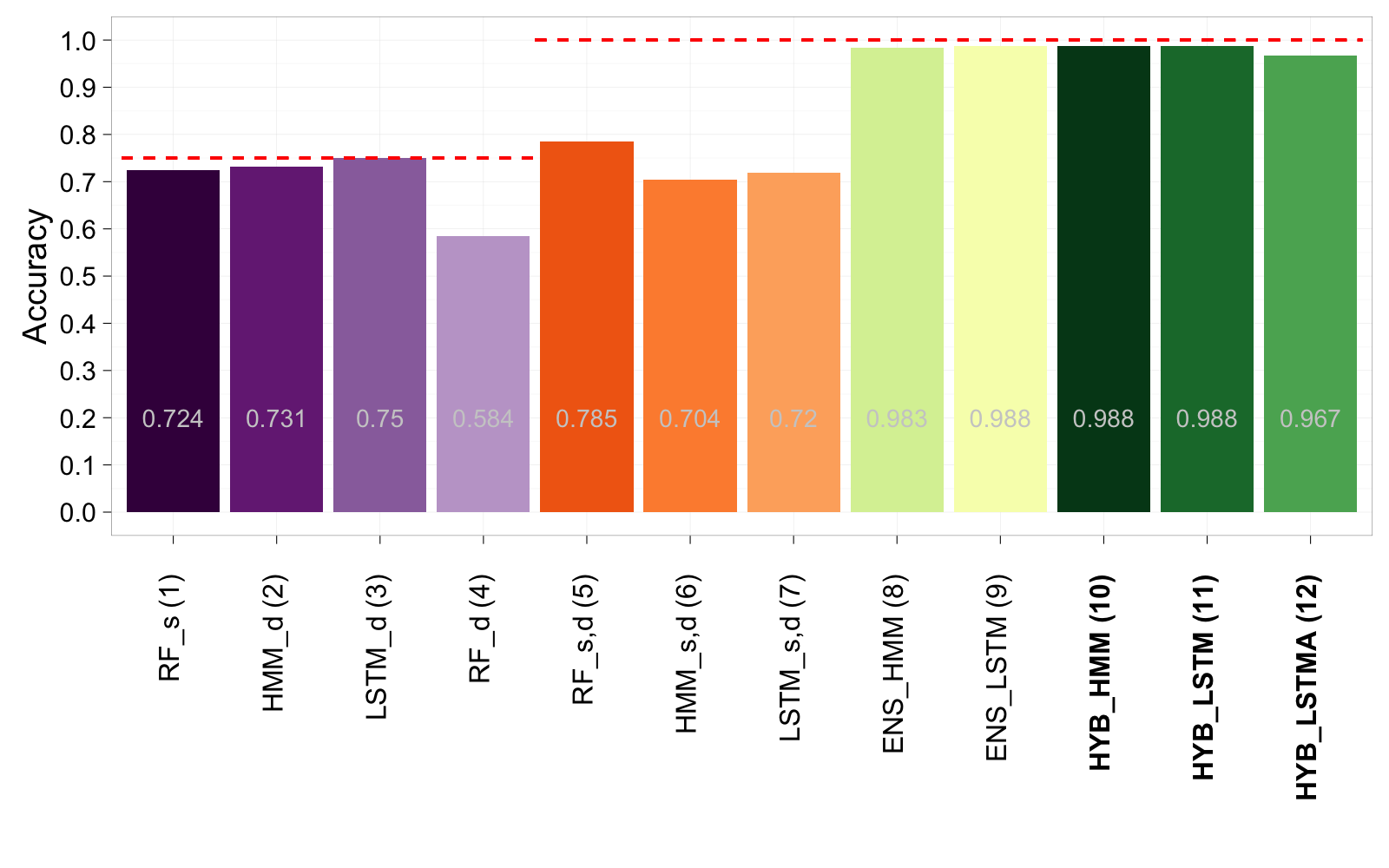}
    \caption{Performance on the synthetic dataset. The vertical dashed lines show the maximal achievable level of performance.}
    \label{fig:synthetic-performance-all}
\end{figure}

In this section we report accuracies achieved by all of the approaches on synthetic and real-life datasets and discuss our findings.
\ \\
\begin{figure*}[tp]
    \centering
    \includegraphics[width=1\linewidth]{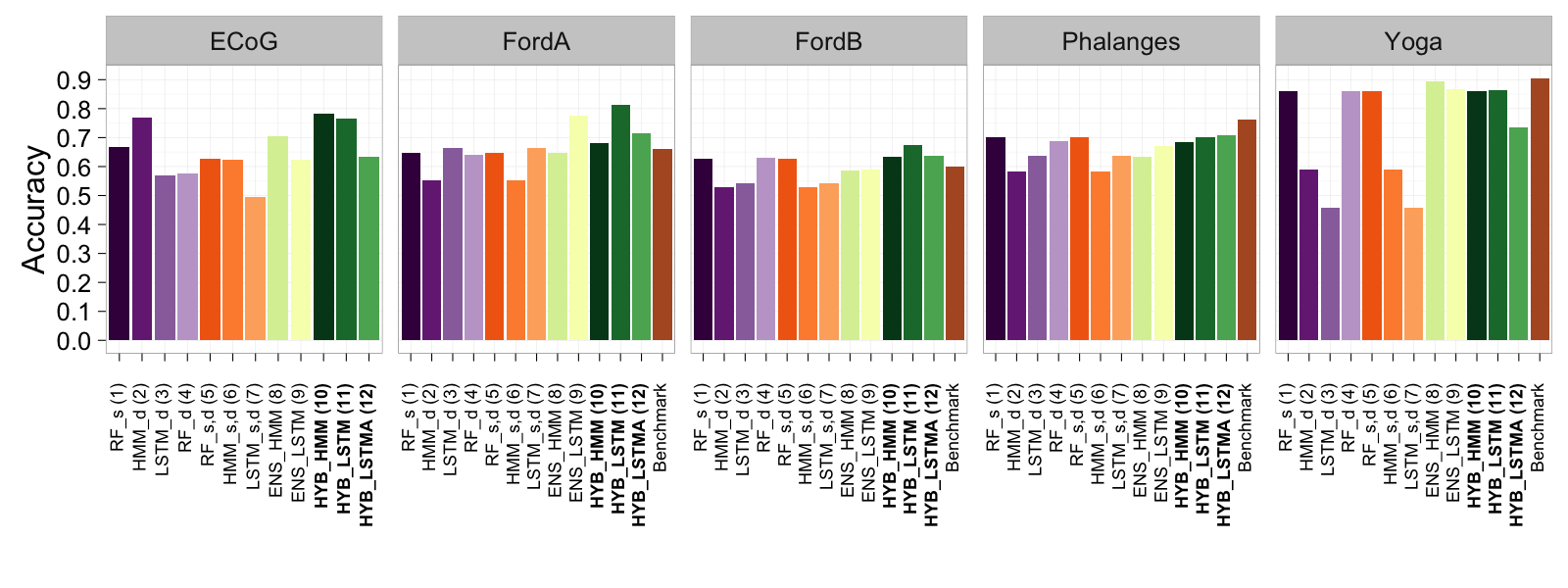}
    \caption{Performance on the real-life datasets.}
    \label{fig:acc_datasets}
\end{figure*}
\paragraph{Insights from synthetic ARMA Dataset} As can be seen from the results on the synthetic dataset (Figure \ref{fig:synthetic-performance-all}) RF is far from good when dealing with dynamic data, thus it confirms the intuition that models designed to work with dynamic data have merit. Stand-alone models on bimodal data (models (5)--(7), see section~\ref{sec:single-on-bi}) fail to capture the information from the both sources. This observation is also generally true for the real-life datasets (see Figure \ref{fig:acc_datasets}). Both the ensemble and hybrid methods achieve almost perfect accuracy, and thus are good at extracting information from temporal and static sources simultaneously.

Next, we investigate various dataset characteristics with respect to the methods' performance by generating datasets with varying size, sequence length ($l_d$), number of static ($n_s$) and dynamic features ($n_d$). The resulting heatmap is shown in Figure \ref{fig:heatmap}. Note that the ratio on the horizontal axis is represented in discrete form and the intervals are not equal due to a limited variation of the parameters. 

There are a few observations that may be of interest for practitioners:
\begin{itemize}
    \item Both ensemble and hybrid HMM (models $\mathit{ENS_{HMM}}\ (8)$, $\mathit{HYB_{HMM}}\ (10)$) show superior performance on longer sequences and when the number of dynamic and static features is balanced.
    \item Hybrid and ensemble models based on LSTM ($\mathit{ENS_{LSTM}}\ (9)$, $\mathit{HYB_{LSTM}}\ (11)$) are less affected by the length of the sequence and the ratio between static and dynamic features. They perform very well across the whole range of those parameters, except for very short sequences.
    \item All methods show lower performance on the datasets with very short sequences and on the datasets with high imbalance towards dynamic features. The decrease is visible when the proportion of dynamic features reaches $80\%$.
    \item Noteworthy is also the overall similarity between the patterns of ensembles and hybrids, though latter have a bit higher accuracy.
    \item Hybrid model $\mathit{HYB_{LSTMA}}\ (12)$ with LSTM activations seems to perform well only for a bigger dataset size, longer sequences and with only a few dynamic features.
\end{itemize}
\medskip
\medskip
\paragraph{Results on real-life datasets} The results across all real-life datasets are shown in Figure \ref{fig:acc_datasets}. In general, every dataset has its own specifics and thus, no single method performs best on all of them. For example, on the ECoG dataset the model $\mathit{HMM_d}\ (2)$ is almost as good as hybrid methods in spite of showing poor performance on all the other datasets. On the Yoga and Phalanges datasets Random Forest on the static features performs as well as the hybrid and ensemble methods. One possible reason why on some datasets combination of dynamic and static features does not give higher performance could be the absence of temporal dynamics in the data. In such case the use of temporal models is irrelevant. The degree of temporal connection can be estimated by comparing stand-alone models $\mathit{HMM_d}\ (2)$ and $\mathit{LSTM_d}\ (3)$ with $\mathit{RF_{d}}\ (4)$. If Random Forest performs on dynamic data better or similar to HMM and LSTM, then the temporal aspect is not present (or temporal models fail to grasp it). 

Hybrid model $\mathit{HYB_{LSTM}}\ (11)$ improved on the best result from the literature on the datasets FordA and FordB. On the ECoG dataset both hybrids $\mathit{HYB_{HMM}}\ (10)$ and $\mathit{HYB_{LSTM}}\ (11)$ outperformed the other methods. Moreover, despite the fact that the hybrid methods on Phalanges and Yoga datasets do not beat the accuracy from the literature, in all of the datasets they are either the best or very close to the best results. 

It is also interesting to notice that the observations obtained from the analysis of synthetic data are in line with the performance on the real datasets. Namely, if a dataset has fewer samples or the sequences are rather short (as in the case of Yoga and Phalanges datasets), then the hybrid approach does not provide a performance boost. However, if a dataset is large and has long sequences as in the case of ECoG, FordA and FordB (see table \ref{table:datasets} for the dataset characteristics), then hybrids outperform other methods.

%
%
\section{Conclusion}
We propose and explore an unorthodox way of combining dynamic and static features in order to make it possible for a classification model to capture temporal dynamics and static information simultaneously. Previous approaches to this problem relied on ensemble methods where different models operate on different data modalities and their predictions are combined. The hybrid approach we propose goes to a lower level and explores the possibility of combining models not at the level of predictions, but earlier --- it concatenates static features with either predictions of generative models, ratios of class probabilities, or, when possible, with inner representation of the data provided by a generative model. We demonstrate that this approach outperforms other approaches and report results on several public datasets. Additionally we explore the behaviour of 12 different models on synthetic data and describe how performance depends on such properties of a dataset as the number of samples, number of static and dynamic features and the length of the sequence, providing the guidelines for practitioners.

%
\section*{Acknowledgements}
We would like to thank Marlon Dumas, Raul Vicente, Tambet Matiisen, Elena Sügis and Dmytro Fishman for their comments on the manuscript. This research was supported by ERDF via the STACC Competence Centre and the Estonian Research Council via the grant PUT438. All experiments were carried out in the High Performance Computing Center of University of Tartu.

%
%
\bibliographystyle{IEEEtran}
\bibliography{IEEEabrv,statdyn}
\medskip

%
%
\appendix
\section{Source Code}
The implementation of all the methods described in the paper and the code for the exploratory analysis is available at the public repository at \url{https://github.com/annitrolla/Generative-Models-in-Classification}.

\end{document}